\documentclass{article}
\usepackage{spconf,amsmath,graphicx,hyperref}
\usepackage{amssymb} 
\usepackage{booktabs}
\usepackage{float}          
\usepackage[H]{algorithm}   
\usepackage{algpseudocode}  
\usepackage{afterpage}
\usepackage{xcolor}
\usepackage{makecell} 
\usepackage{booktabs}
\usepackage{tabularx}
\usepackage{multirow}
\usepackage[table]{xcolor} 
\usepackage{colortbl}     

\title{Head-Aware Visual Cropping: Enhancing Fine-Grained VQA with Attention-Guided Subimage
}
\name{Author(s) Name(s)\thanks{Thanks to XYZ agency for funding.}}
\address{Author Affiliation(s)}
\name{Junfei Xie$^{1,2\ast}$, Peng Pan$^{1,2\ast}$, Xulong Zhang$^{1\dagger}$
\thanks{$^{\star}$ These authors contributed equally to this work.}%
\thanks{$^{\dagger}$ Corresponding author.}
\thanks{Supported by Shenzhen-Hong Kong Joint Funding Project (Category A) under grant No. SGDX20240115103359001.}
}

\address{$^1$Ping An Technology (Shenzhen) Co., Ltd., Shenzhen, China \\
         $^2$University of Science and Technology of China, Hefei, China
        }

\begin{document}
\ninept
\maketitle
\begin{abstract}
Multimodal Large Language Models (MLLMs) show strong performance in Visual Question Answering (VQA) but remain limited in fine-grained reasoning due to low-resolution inputs and noisy attention aggregation. We propose \textbf{Head Aware Visual Cropping (HAVC)}, a training-free method that improves visual grounding by leveraging a selectively refined subset of attention heads. HAVC first filters heads through an OCR-based diagnostic task, ensuring that only those with genuine grounding ability are retained. At inference, these heads are further refined using spatial entropy for stronger spatial concentration and gradient sensitivity for predictive contribution. The fused signals produce a reliable Visual Cropping Guidance Map, which highlights the most task-relevant region and guides the cropping of a subimage subsequently provided to the MLLM together with the image-question pair. Extensive experiments on multiple fine-grained VQA benchmarks demonstrate that HAVC consistently outperforms state-of-the-art cropping strategies, achieving more precise localization, stronger visual grounding, providing a simple yet effective strategy for enhancing precision in MLLMs.

\end{abstract}
\section{Introduction}
\label{sec:intro}
Recent research on Multimodal Large Language Models (MLLMs) \cite{liu2023visual, dai2023instructblip,chen2024internvl,zhou2022learning,wang2024qwen2} has achieved remarkable progress in Visual Question Answering (VQA)\cite{first2,lu2026vista}, demonstrating strong cross-modal perception and reasoning abilities. Despite these advances, current models remain fundamentally constrained in their ability to handle high-resolution inputs\cite{DC2Wang2025,zhang-mllms2025}. Architectures such as InstructBLIP \cite{dai2023instructblip} and LLaVA \cite{liu2023visual} are typically pre-trained with fixed input sizes, which severely restricts their capacity to capture fine-grained details. This loss of detail undermines the model’s ability to ground its predictions, leading to performance degradation on fine-grained VQA tasks\cite{WuV*-2024,li2023evaluating,schwenk2022okvqa,singh2019towards}.

To address this challenge, recent research has centered on two mainstream paradigms: Static Input Enhancement and Active Visual Exploration. The former seeks to preserve fine-grained details directly within the initial visual input, most commonly through patching and feature integration\cite{LiMonkey-2024, DBLP:conf/iclr/BergnerLM23, Zhao-zoomvqa-2023, yin2022vit,shi2025scaling,zhao2025RUNA}. For instance, AnyRes\cite{li2024llava} divides a high-resolution image into multiple patches and encodes them together with a downsampled global view, enabling a comprehensive single-pass analysis. While effective for retaining details, this approach inevitably increases the number of visual tokens, leading to higher computational overhead and limited scalability at very high resolutions. In contrast, Active Visual Exploration allows models to perform multi-pass analysis, actively searching for query-relevant evidence in a human-like manner. This paradigm has been instantiated in various forms, including cropping salient regions guided by internal attention signals \cite{zhang-mllms2025}, conducting LLM-driven searches to recover missing information and integrate it into a visual working memory \cite{WuV*-2024}, and navigating hierarchical image structures to simulate zooming \cite{Zhao-zoomvqa-2023}. Despite these advances, its effectiveness is limited by the guidance mechanism, which typically aggregates attention from all heads within a single layer \cite{zhang-mllms2025, kang2025your}. This practice conflates informative and uninformative signals, impairing the precision of localization. Recent studies reveal that only a sparse subset of specialized visual heads are causally responsible for visual grounding, while the majority cause noise, undermining the reliability of attention maps \cite{SparseMM2025, retrive_head2025, Devils_in_Middle_Layers2025}. These findings highlight a critical gap in current approaches and motivate the design of more selective and interpretable guidance strategies for fine-grained VQA.

Building on the insight that only a sparse subset of specialized visual heads are truly effective, our work introduces a two-stage methodology to address the noise problem from indiscriminate head aggregation. In the first stage, we identify expert visual heads via an OCR-based diagnostic task\cite{SparseMM2025}: by aligning attention peaks with ground-truth text regions, we derive a projection-based score that quantifies each head’s ability to achieve fine-grained visual grounding. This filtering ensures that only heads with genuine visual grounding capability are preserved. In the second stage, at inference, these candidate heads are further refined through two complementary signals. The first signal, spatial entropy, evaluates the compactness of attention distributions and keep heads with stronger spatial concentration. The second signal, gradient sensitivity, measures predictive contribution by quantifying how much increasing a head’s attention improves answer confidence. After normalization and weighted fusion of these two signals, the top-K heads are aggregated to generate a Visual Cropping Guidance Map, which highlights the most task-relevant region of the image. By cropping and providing a subimage to the MLLM together with the original input, our method directs the model’s reasoning toward evidence-rich areas, leading to more precise grounding and superior performance on fine-grained VQA tasks. Our main contributions are summarized as follows:

\begin{itemize}
\item[$\bullet$] We propose a novel cropping framework that identifies a sparse set of expert visual heads, mitigating noise from indiscriminate aggregation and enabling more interpretable and precise visual grounding.

\item[$\bullet$] We design a dual-branch refinement strategy that fuses complementary signals, spatial entropy for stronger spatial concentration and gradient sensitivity for predictive contribution, to construct a reliable Visual Cropping Guidance Map.

\item[$\bullet$] We conduct extensive experiments on multiple fine-grained VQA benchmarks, demonstrating that our mechanism achieves consistent and significant improvements.
\end{itemize}

\section{METHOD}
\label{sec:method}

Our approach first identifies expert visual heads through an OCR-based diagnostic task, ensuring that only heads with strong visual grounding ability are preserved. At inference, these heads are further refined by spatial-entropy filtering and gradient sensitivity scoring, which measure attention compactness and predictive contribution. The resulting scores are normalized, fused, and used to select the top-$K$ heads, whose attention maps are aggregated into a Visual Cropping Guidance Map that highlights the most task-relevant image region for cropping.

\subsection{Identifying Expert Visual Heads}
\label{ssec:stage1}

To identify which attention heads genuinely contribute to visual grounding, we introduce an OCR-based method and compute a visual score measuring their alignment with ground-truth regions.

Let the multimodal input sequence be denoted as $\Omega = \{x_i\}_{i=1}^{L}$, which contains textual and visual tokens, as well as special tokens, the sequence length is denoted as $L = |\Omega|$. We exclude the latter from the search space and denote the remaining valid tokens by $\Omega' \subseteq \Omega$. The subset of visual tokens is denoted as $V \subseteq \Omega'$. For each output token $y_i$ ($i=1,\dots,N$), we construct a visual grounding mask $m_i \in \{0,1\}^{|\Omega|}$, where the entries corresponding to the aligned visual region are set to 1, and all others are set to 0. The region size is $\|m_i\|_1$, i.e., the number of visual tokens covered by $m_i$. 

\begin{figure}[!htbp]
    \centering
    \includegraphics[width=\columnwidth]{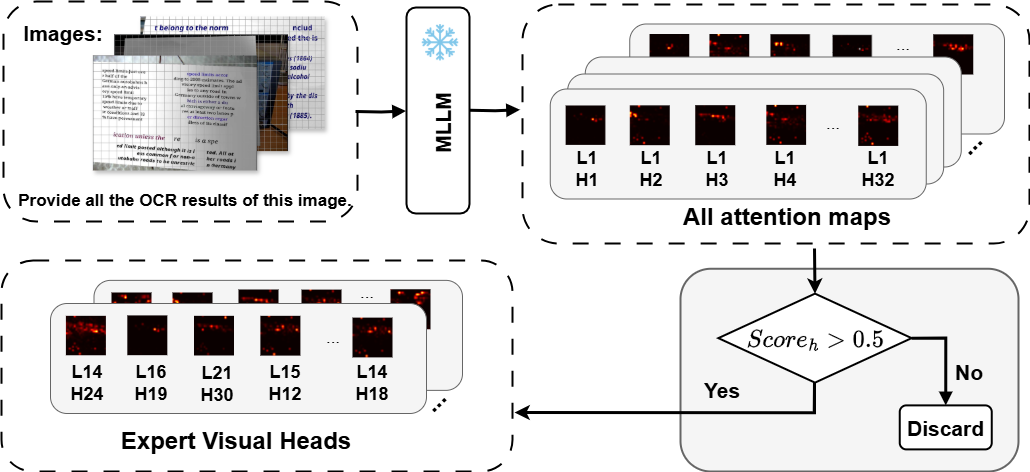} 
    \caption{Overview of identifying Expert Visual Heads. Through an OCR-based diagnostic task, the model generates all attention maps, and heads with a visual grounding score above 0.5 are retained as expert visual heads for further refinement.
}
    \label{主图}
\end{figure}

For any given head $h$, when predicting $y_i$, let $A_h^{(i)} \in\mathbb{R}^{|\Omega|}$ denote the attention distribution over all input tokens. We define the peak index as
\begin{equation}
\label{eq:peak_index}
j^{*} = \arg\max_{j \in \Omega'} A_h^{(i)}[j],
\end{equation}
where $j$ indexes the input tokens in $\Omega'$ excluding special tokens such as \texttt{[BOS]} or \texttt{[IMG\_CLS]}. The peak vector $p_h^{(i)}$ is then defined as
\begin{equation}
\label{eq:peak_vector}
p_h^{i}[j] =
\begin{cases}
1, & j = j^* \\
0, & \text{otherwise}
\end{cases}
\end{equation}

We then measure whether the attention peak falls inside the ground-truth visual region by computing the following score:
\begin{equation}
\label{eq:proj_score}
ProjScore_h^{i}= 
\frac{\langle p_h^{i},\, m_i \rangle}{\|m_i\|_1},
\end{equation}
This equals $1/\|m_i\|_1$ if the peak index lies within the correct visual region of $y_i$, and $0$ otherwise. The normalization by $\|m_i\|_1$ accounts for region size, rewarding heads that focus on small, fine-grained visual areas.

Finally, the visual grounding score of the $h$-th head is defined as the average projection score across all output tokens:
\begin{equation}
\label{eq:head_score}
Score_h = \frac{1}{N} 
\sum_{i=1}^{N} {ProjScore_h^{i}}.
\end{equation}

As illustrated in Fig.~\ref{主图}, the visual head filtering process begins with a diagnostic OCR setup, where the model is prompted with images containing text. During generation, whenever the predicted token matches the ground-truth text, we record the head’s peak position and update its projection score. Repeating this process across 1,000 OCR
images from the Synthdog dataset yields an accumulated score matrix $\mathbf{S}$. We then apply min–max normalization to each element of the score matrix. Heads whose normalized scores exceed 0.5 are retained as expert visual heads, highlighted in the final matrix.

\subsection{Visual Cropping Guidance Map Generation}
\label{ssec:stage2}

\begin{figure}[!htbp]
    \centering
    \includegraphics[width=\columnwidth]{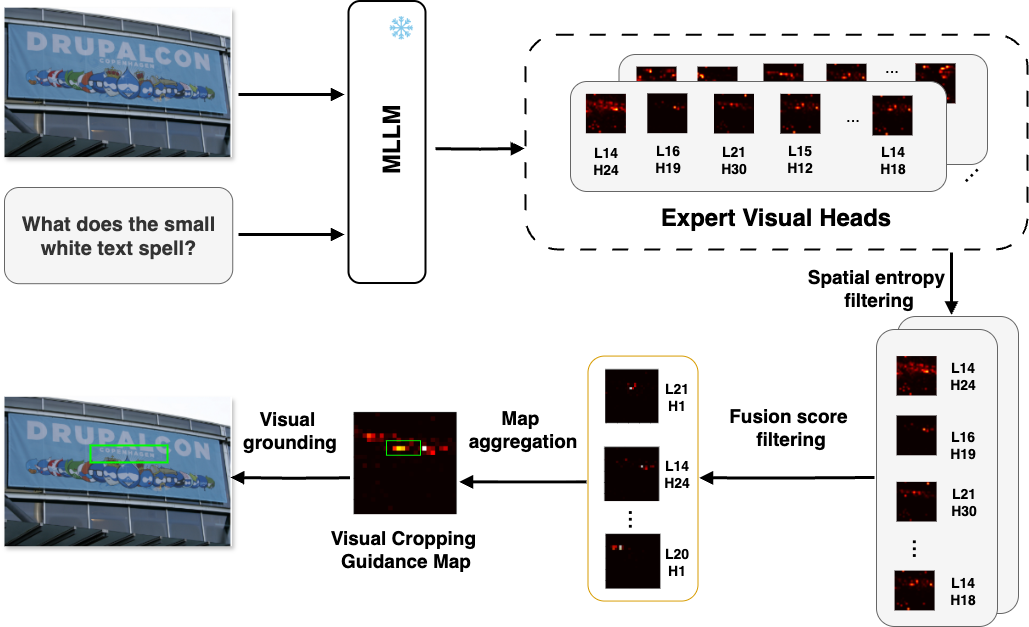} 
    \vspace{-3mm}
    \caption{Overview of the Visual Cropping Guidance Map Generation Process. Expert visual heads are further refined by spatial entropy and fusion score filtering at inference time. Their attention maps are then aggregated to produce a guidance map, which highlights task-relevant regions and directs the final visual grounding.
    \label{futu}
}
\end{figure}

After filtering the expert visual heads (Sec.~\ref{ssec:stage1}), as illustrated in Fig.~\ref{futu}, we further refine them at inference to obtain a Visual Cropping Guidance Map. The process consists of three steps: (i) spatial-entropy filtering to keep heads with stronger spatial concentration, (ii) gradient-based scoring to measure each head’s predictive contribution, and (iii) score fusion and map aggregation to construct the final guidance map.

For each candidate head $h$, we extract the attention vector $\mathbf{a}_h \in \mathbb{R}^{N_{\text{img}}}$, which represents the distribution of the current decoding query over all visual tokens. The attention vector $\mathbf{a}_h$ is reshaped into a 2D attention map $\mathbf{A}_h \in \mathbb{R}^{N_p \times N_p}$, following the row-major order of image patches, where $N_p = \sqrt{N_{\text{img}}}$.

\begin{table*}[!h]
\centering
\caption{Accuracy comparison of Vanilla, ViCrop, and our HAVC on LLaVA-1.5 and InstructBLIP across six benchmarks with different signal types. The best results are bolded, and the second-best are underlined.}
\label{main_table}

\begin{tabular*}{\textwidth}{@{\extracolsep{\fill}}lllcccccc}
\toprule
\textbf{Backbone} & \textbf{Method} & \textbf{Signal Type} & \textbf{AOKVQA} & \textbf{POPE} & \textbf{TextVQA} & \textbf{V*} & \textbf{VQAv2} & \textbf{GQA} \\
\midrule[\heavyrulewidth]

\multirow{5}{*}{\textit{LLaVA-1.5}} 
 & Vanilla & --- & 59.71 & 83.90 & 46.88 & 42.93 & 76.31 & 71.40 \\
\cmidrule(l){2-9} 

 & \multirow{3}{*}{ViCrop} 
   & grad-att & 60.40 & \underline{84.90} & \underline{56.15} & 46.59 & \underline{77.25} & 72.10 \\
 & & rel-att  & \textbf{61.21} & 84.70 & 54.84 & 47.64 & 77.05 & \underline{72.30} \\
 & & pure-att & 60.74 & 84.00 & 51.42 & \underline{49.00} & 77.03 & 71.30 \\
\cmidrule(l){2-9}

 & HAVC(ours) & grad+entropy & \underline{60.95} & \textbf{85.80} & \textbf{57.60} & \textbf{49.73} & \textbf{77.78} & \textbf{72.50} \\

\midrule[\heavyrulewidth]

\multirow{5}{*}{\textit{InstructBLIP}} 
 & Vanilla & --- & 60.65 & 81.4 & 35.69 & \textbf{37.69} & 74.41 & 56.1 \\
\cmidrule(l){2-9}

 & \multirow{3}{*}{ViCrop}
   & grad-att & 59.30 & 82.00 & 38.19 & 34.55 & 74.52 & 56.30 \\
 & & rel-att & 60.97 & 82.40 & \underline{41.10} & 37.17 & 74.47 & 57.40 \\
 & & pure-grad & \textbf{61.24} & \textbf{82.90} & 36.37 & 35.07 & \underline{74.61} & \underline{57.80} \\
\cmidrule(l){2-9}

 & HAVC(ours) & grad+entropy & \underline{61.04} & \underline{82.70} & \textbf{41.82} & \underline{37.17} & \textbf{76.37} & \textbf{57.90} \\

\bottomrule
\end{tabular*}
\end{table*}

\textbf{Spatial entropy filtering.}  
For spatial entropy computation, $\mathbf{A}_h$ is normalized to $[0,1]$, thresholded by Otsu’s method to obtain a binary mask $\mathbf{b}_h$, and decomposed into $C_h$ connected components with centroids $\{\boldsymbol{c}_r\}_{r=1}^{C_h}$. Let $d_{\max}=\sqrt{N_p^2+N_p^2}$ be the diagonal length of the map, and $\bar{d}_h$ the mean pairwise centroid distance. We define the spatial entropy of head $h$ as
\begin{equation}
\label{eq:entropy}
E_h = \min\!\Bigg(\lambda_c \cdot (C_h-1) + \lambda_d \cdot \frac{\bar{d}_h}{d_{\max}}, \;1\Bigg),
\end{equation}
where the first term penalizes multiple scattered components, and the second measures centroid dispersion. A smaller $E_h$ indicates stronger spatial concentration. Only heads with $E_h$ below a preset threshold (0.3 in our implementation) are retained, where our parameter settings follow \cite{kang2025your}.

\textbf{Gradient sensitivity scoring.}
To quantify each head’s predictive contribution, we measure how much its increasing visual attention would affect the model’s predicted probability. Let $y^\ast$ be the predicted token and $p(y^\ast)$ its probability. For each head $h$, we compute the gradient sensitivity:
\begin{equation}
\label{eq:sens}
\mathbf S_h^\text{sens} = \frac{\partial \log p(y^\ast)}{\partial \mathbf{a}_h}, 
\end{equation}
and we retain only positive gradients. Weighting them by the original attention via an inner product yields the gradient score
\begin{equation}
G_h = \langle \mathbf{a}_h, \max(0, \mathbf S_h^\text{sens}) \rangle ,
\end{equation}
A large $G_h$ suggests that the head attends to useful visual tokens where increasing attention would improve the prediction confidence.

\textbf{Score fusion and map aggregation.}  
Since a smaller $E_h$ indicates stronger spatial concentration, we replace it with $1-E_h$ so that larger values correspond to better concentration.
\begin{equation}
S_h = \alpha \cdot N(1 - E_h) + (1-\alpha)\cdot N(G_h),
\end{equation}
where $N(\cdot)$ denotes min–max normalization across heads and $\alpha$ serves as a weighting factor controlling the balance between spatial concentration and predictive contribution. Heads are ranked by $S_h$ and the top-K are retained.

For each selected head, the fusion weights are obtained via a temperature-scaled softmax:
\begin{equation}
w_h = \frac{\exp(S_h/\tau)}{\sum_{j=1}^{K} \exp(S_j / \tau)},
\end{equation}
where $\tau$ is a temperature parameter controlling the sharpness of the distribution. 
Then the Visual Cropping Guidance Map is the weighted sum:
\begin{equation}
\mathbf{M}_{\text{final}} = \sum_{h=1}^{K} w_h \cdot \mathbf{A}_h.
\end{equation}

The resulting $\mathbf{M}_{\text{final}}$ highlights the most task-relevant region. 
Following \cite{zhang-mllms2025}, we extract a bounding box from $\mathbf{M}_{\text{final}}$ for visual cropping and feed the cropped subimage together with the original image into the MLLM.

\section{EXPERIMENT}
\label{sec:experiment}

\subsection{Experimental Setup}
\label{ssec:setup}
Our experiments are conducted on the LLaVA-1.5 (Vicuna-7B)\cite{Liu-llava1.5-2024} and InstructBLIP (Vicuna-7B)\cite{dai2023instructblip}. We compare our method (HAVC) with two baselines. The first is the original MLLM, which does not perform any visual cropping. The second is derived from ViCrop\cite{zhang-mllms2025}, a state-of-the-art, training-free cropping method, and includes its three mainstream approaches: rel-att, grad-att, and pure-grad. To comprehensively evaluate model performance, we adopted four fine-grained VQA benchmarks (AOKVQA\cite{schwenk2022okvqa}, POPE\cite{li2023evaluating}, TextVQA\cite{singh2019towards}, V*\cite{WuV*-2024}) and two general benchmarks (VQAv2\cite{goyal2017making}, GQA\cite{hudson2019gqa}), and uniformly used Accuracy as the evaluation metric.

\subsection{Main Results}
\label{ssec:main_results}
Detailed quantitative results on two backbones are reported in Table~\ref{main_table}. Overall, HAVC consistently achieves superior or highly competitive performance across diverse benchmarks, demonstrating strong robustness and effectiveness under the training-free setting.

On LLaVA-1.5, HAVC achieves the best results on five out of six benchmarks, including POPE (85.80\%), TextVQA (57.60\%), V* (49.73\%), VQAv2 (77.78\%), and GQA (72.50\%). Compared with the Vanilla baseline, HAVC brings substantial improvements, especially on challenging datasets, e.g., from 46.88\% to 57.60\% on TextVQA and from 42.93\% to 49.73\% on V*. On AOKVQA, HAVC remains highly competitive (60.95\%), closely trailing the best ViCrop (rel-att) variant (61.21\%). Compared with ViCrop using different attention signals, HAVC consistently yields stronger performance on most benchmarks, indicating the benefit of jointly selecting expert heads and fusing complementary signals.

On InstructBLIP, HAVC similarly improves over both Vanilla and ViCrop on most benchmarks. It achieves the best performance on TextVQA (41.82\%), VQAv2 (76.37\%), and GQA (57.90\%), while remaining highly competitive on AOKVQA and POPE (61.04\% and 82.70\%, respectively). On V*, HAVC matches the strongest ViCrop result (37.17\%) and is slightly below the Vanilla baseline (37.69\%), suggesting that this benchmark is less sensitive to cropping strategies on this backbone. Overall, the consistent gains across backbones and datasets confirm the robustness and general applicability of HAVC in diverse visual reasoning scenarios.

\subsection{Ablation Studies}
\label{ssec:ablation}

To validate our key design choices, we conducted a series of ablation studies on the TextVQA benchmark, with the results presented in Table~\ref{ablation_table}. Without cropping, the baseline (No-Crop) achieves 46.36\% accuracy. Random cropping further degrades performance, indicating that naive spatial selection is ineffective. Using all attention heads for cropping (All-Heads) yields only marginal improvements, suggesting that indiscriminate head aggregation introduces substantial noise.
In contrast, filtering expert heads leads to a substantial performance gain. Using only OCR-filtered heads with uniform fusion (Filter Heads) improves accuracy from 47.87\% to 56.52\%, confirming that accurate head selection is the most critical component.
We further evaluate the proposed dual-branch fusion. Incorporating the entropy branch or gradient branch individually yields consistent improvements, and combining both branches achieves the best performance. Our final model reaches 57.60\% accuracy, 68.24\% F1-Score, and 66.59\% precision, demonstrating the effectiveness of both head filtering and dual-branch fusion.

\begin{table}[!h]
\centering
\vspace{-3mm}
\caption{Ablation studies on the TextVQA benchmark to validate our design choices for the heads filtering and fusion mechanisms.}
\label{ablation_table}
\begin{tabularx}{\columnwidth}{Xccc}
\toprule
\textbf{Method Configuration} & \textbf{Acc. ↑} & \textbf{F1-Score ↑} & \textbf{Precision ↑} \\
\midrule
No-Crop & 46.36 & 57.15 & 55.44 \\
\midrule
Random-Crop & 45.95 & 55.80 & 54.53 \\
All-Heads & 47.87 & 58.21 & 56.67 \\
Filter Heads & 56.52 & 67.52 & 65.85 \\
Filter Heads + entropy & 56.96 & 67.74 & 66.10 \\
Filter Heads + gradient & 57.29 & 67.86 & 66.25 \\
\midrule
\textbf{HAVC (Ours)} & \textbf{57.60} & \textbf{68.24} & \textbf{66.59} \\
\bottomrule
\end{tabularx}
\end{table}

\begin{figure}[!t]
    \centering
    \includegraphics[width=\columnwidth]{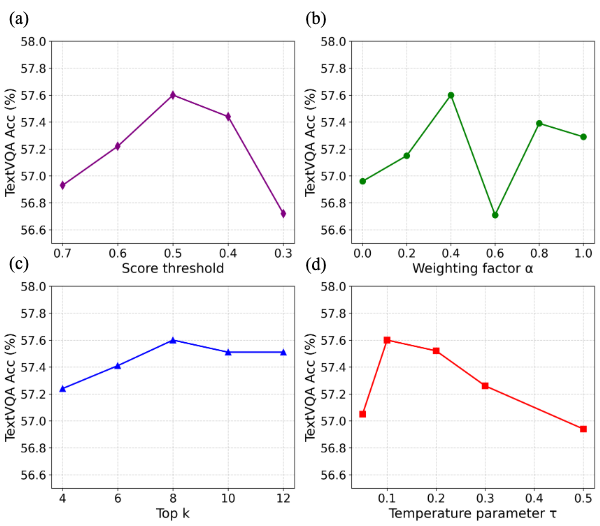} 
    \vspace{-5mm}
    \caption{Sensitivity analysis of key parameters in our HAVC framework, evaluated on the TextVQA dataset. 
(a) Effect of score threshold for expert head filtering. 
(b) Effect of weighting factor $\alpha$ for balancing spatial entropy and gradient sensitivity. 
(c) Effect of the number of selected heads K. 
(d) Effect of temperature parameter $\tau$ in the softmax weighting. 
}
    \label{sensitivity}
\end{figure}

Second, to establish a robust and reproducible configuration, we optimized the key hyperparameters for HAVC through a sensitivity analysis on the TextVQA dataset, as shown in Fig.~\ref{sensitivity}. The results show that our method is robust across a wide range of parameter values, with optimal performance achieved around a score threshold of 0.5, $\alpha=0.4$, $K=8$, and $\tau=0.1$.

\begin{figure}[!t]
    \centering
    \vspace{-1mm}
    \includegraphics[width=\columnwidth]{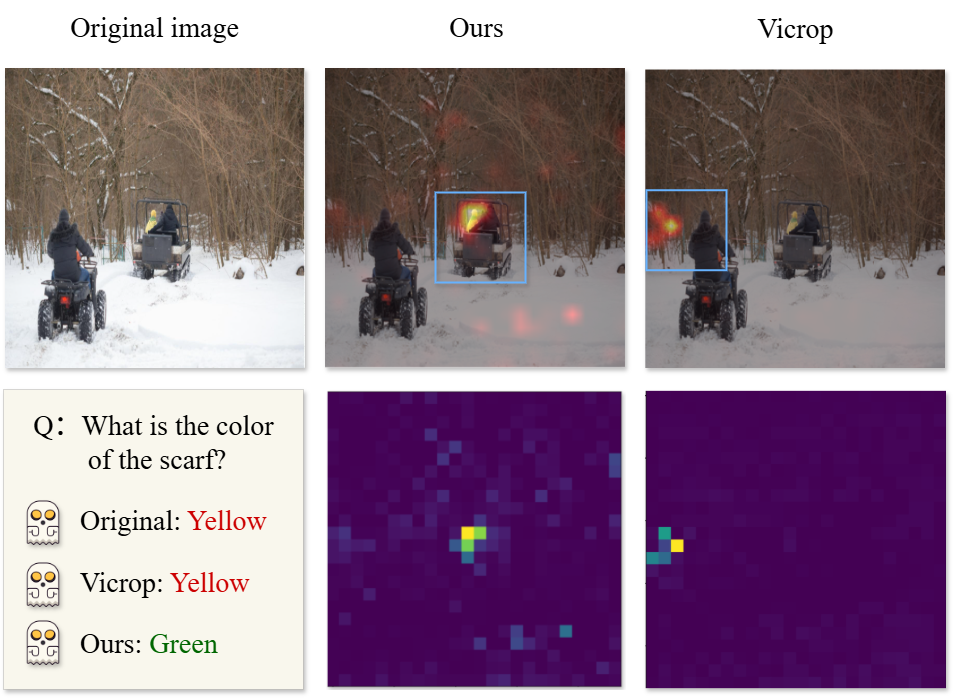} 
    \vspace{-3mm}
    \caption{Qualitative comparison for the question ``What is the color of the scarf?". Our method correctly localizes the target and provides the right answer, while the baselines (Original, ViCrop) fail.}
    \vspace{-3mm}
    \label{keshihua}
\end{figure}

\subsection{Qualitative Analysis}
\label{ssec:qualitative}

Fig.~\ref{keshihua} provides a qualitative comparison for the question, ``What is the color of the scarf?". Both the original MLLM and the ViCrop fail, incorrectly predicting ``Yellow". This error occurs because their attention is distracted by other yellow objects in the scene, such as the girl's hat and clothes, causing them to miss the actual target. In contrast, our HAVC method (Ours) successfully overcomes this challenge. Its guidance map accurately localizes the green scarf, ignoring the surrounding distractors. This precise cropping guides the MLLM to provide the correct answer, ``Green", demonstrating the reliability of HAVC in fine-grained reasoning.



\section{CONCLUSION}
\label{sec:copyright}

In this work, we introduced HAVC, a training-free framework that enhances fine-grained VQA by selectively leveraging expert visual heads. Through an OCR-based diagnostic task and dual-branch refinement, HAVC generates reliable cropping guidance map that highlights task-relevant regions. Extensive experiments demonstrate consistent improvements over existing cropping strategies.


\bibliographystyle{IEEEbib}
\bibliography{strings,refs}

\end{document}